\documentclass{article}
\usepackage{ijcai20}

\usepackage{times}
\usepackage{soul}
\usepackage{url}
\usepackage[hidelinks]{hyperref}
\usepackage[utf8]{inputenc}
\usepackage[small]{caption}
\usepackage{graphicx}
\usepackage{amsmath}
\usepackage{amsthm}
\usepackage{booktabs}
\usepackage{algorithm}
\usepackage{algorithmic}
\usepackage{amsmath}
\usepackage{amssymb}
\usepackage{color}
\usepackage{booktabs}
\usepackage{multirow,array}
\usepackage{float}
\usepackage{bm}
\newcommand\myacc[2] { $#1 \scriptstyle\small \pm #2$}
\newcommand\maxacc[2] { $\mathbf{#1 \scriptstyle\small \pm #2}$}
\newcommand{\paraheading}[1]{\vspace{0.6em} \noindent \textbf{#1} \hspace{0.1em}}

\title{Asymmetric Distribution Measure for Few-shot Learning}

\author{Wenbin Li\textsuperscript{$1$},\quad Lei Wang\textsuperscript{$2$},\quad Jing Huo\textsuperscript{$1$},\quad Yinghuan Shi\textsuperscript{$1$},\quad Yang Gao\textsuperscript{$1$},\quad Jiebo Luo\textsuperscript{$3$} \\
\textsuperscript{$1$}Nanjing University, China,\quad
\textsuperscript{$2$}University of Wollongong, Australia,\\
\textsuperscript{$4$}University of Rochester, USA\\
\emails
liwenbin.nju@gmail.com,
leiw@uow.edu.au,
\{huojing, syh, gaoy\}@nju.edu.cn,
jluo@cs.rochester.edu
}

\begin{document}

\maketitle

\begin{abstract}
The core idea of metric-based few-shot image classification is to directly measure the relations between query images and support classes to learn transferable feature embeddings. Previous work mainly focuses on image-level feature representations, which actually cannot effectively estimate a class's distribution due to the scarcity of samples. Some recent work shows that local descriptor based representations can achieve richer representations than image-level based representations. However, such works are still based on a less effective instance-level metric, especially a symmetric metric, to measure the relations between query images and support classes. Given the natural asymmetric relation between a query image and a support class, we argue that an asymmetric measure is more suitable for metric-based few-shot learning. To that end, we propose a novel \textit{Asymmetric Distribution Measure (ADM)} network for few-shot learning by calculating a joint local and global asymmetric measure between two multivariate local distributions of queries and classes. Moreover, a task-aware \textit{Contrastive Measure Strategy (CMS)} is proposed to further enhance the measure function. On popular \textit{mini}ImageNet and \textit{tiered}ImageNet, we achieve $3.02\%$ and $1.56\%$ gains over the state-of-the-art method on the $5$-way $1$-shot task, respectively, validating our innovative design of asymmetric distribution measures for few-shot learning.
\end{abstract}

\section{Introduction}

Few-shot learning for image classification has gained considerable attention in recent years~\cite{vinyals2016matching,finn2017model,sung2018learning,lee2019meta}, which attempts to learn a classifier with good generalization capacity for new unseen classes with only a few samples. Because of the scarcity of data, it is almost impossible to directly train a conventional supervised model (\emph{e.g.,} a convolutional neural network) from scratch by only using the few available samples. Therefore, transfer learning shall be a natural way to learn transferable knowledge to boost the target few-shot classification. Along this way, a variety of methods have been proposed, which can be roughly divided into three categories: {\it data-augmentation based} methods~~\cite{antoniou2017data,schwartz2018delta,xian2019f}, {\it meta-learning based} methods~\cite{ravi2016optimization,jamal2019task,lee2019meta} and {\it metric-based} methods~\cite{vinyals2016matching,sung2018learning,li2019CovaMNet}. Metric-based few-shot learning methods have achieved significant successes and attracted increasing attention due to their simplicity and effectiveness. In this work, we focus on this kind of methods.

The basic idea of metric-based few-shot learning methods is to learn a transferable deep embedding network by directly measuring the relations between query images and support classes. Thus, two key issues are involved in such a kind of methods, \textit{i.e.,} feature representations and relation measure. For feature representations, traditional methods such as ProtoNet~\cite{snell2017prototypical} and RelationNet~\cite{sung2018learning} generally adopt image-level global feature representations for both query images and support classes. However, due to the scarcity of samples in each class, such image-level global features are not effective in representing the underlying distribution of each class. Recently, CovaMNet~\cite{li2019CovaMNet} and DN4~\cite{li2019DN4} introduce deep local descriptors into few-shot learning and attempt to utilize the distribution of local descriptors to represent each support class, which have been verified to be more effective than using the image-level global features.

On the relation measure, the existing methods including the above methods usually adopt an instance-level metric, where the query image is taken as one single instance (\textit{i.e.,} an image-level feature representation) or a set of instances (\textit{i.e.,} a set of local feature descriptors). For example, in ProtoNet, the Euclidean distance is chosen to calculate the distance between a query instance and the prototype (\emph{i.e.,} the mean vector) of each support class. Also, CovaMNet proposes a covariance metric function to measure a local similarity between each local descriptor of a query image and a support class. Next, it aggregates all the local similarities to obtain a global similarity as the relation between this query image and this class.

However, these existing methods have only considered the distributions of the support classes while neglecting the natural distributions of the query images. Moreover, the instance-level metric they employ can only capture a kind of local relations (\emph{i.e.,} local similarities) between the query images and support classes. We argue that the distributions of the query images are equally important and a \textit{distribution-level measure} shall be designed to capture the global-level relations between the queries and classes. More importantly, we observe that the existing methods usually adopt a symmetric metric function (\emph{i.e.,} $M(a,b)=M(b,a)$) to calculate the symmetric relations between queries and classes. For instance, both the Euclidean distance used in ProtoNet and the cosine similarity adopted in CovaMNet and DN4 are symmetric functions. However, we highlight that there is an asymmetric relation between the query images and a certain class. In particular, when each image is represented by a set of deep local descriptors, the distribution of the descriptors in one query image is only comparable to part of the distribution of the descriptors extracted from a support class. Therefore, we argue that an asymmetric measure is more suitable for the metric-based few-shot learning to capture the asymmetric relations. 

To this end, we develop a novel \textit{Asymmetric Distribution Measure (ADM)} network for metric-based few-shot learning. First, we represent each image as a set of deep local descriptors (instead of a single image-level global feature representation) and consider characterizing both query images and support classes from the perspective of local descriptor based distributions (\textit{i.e.,} mean vector and covariance matrix). Second, we employ an asymmetric Kullback--Leibler (KL) divergence measure to align a query distribution with a support class distribution to capture the global distribution-level asymmetric relations. Third, to further improve the metric space by taking the context of the task into consideration, we propose a task-aware \textit{Contrastive Measure Strategy (CMS)}, which can be used as a plug-in to any measure functions. Finally, inspired by the successful image-to-class measure (an asymmetric measure as a whole) introduced in DN4 which mainly captures the asymmetric relations via individual local descriptor based cosine similarity measures, we combine the whole distribution based KL divergence measure with the image-to-class measure together to simultaneously capture the global and local relations.

The main contributions of this work are as follows: 
 \begin{itemize}
    \item We propose a pure distribution based method for metric-based few-shot learning and show that an asymmetric measure is more suitable for this kind of few-shot learning methods.
    \item We simultaneously combine the global relations (\textit{i.e.,} the KL divergence measure) and the local relations (\textit{i.e.,} the image-to-class measure) together to measure the complete asymmetric distribution relations between queries and classes.
     \item We propose an adaptive fusion strategy to adaptively integrate the global and local relations.
    \item We design a task-aware contrastive measure strategy (CMS) as a plug-in to further enhance the adopted measure functions.
\end{itemize}

\section{Related Work}
We first briefly review the metric-based few-shot learning methods in the literature, and then introduce related work that inspired our work in this paper.

The first metric-based few-shot learning method was proposed in~\cite{koch2015siamese}, which adopted a Siamese neural network to learn transferable and discriminative feature representations. In~\cite{vinyals2016matching}, a Matching Net which directly compares the query images with the support classes was presented, where a subsequently widely used episodic training mechanism was also proposed. After that, \cite{snell2017prototypical} proposed a ProtoNet, which represents a support class by a prototype, \textit{i.e.,} the mean vector of all sample in this class. Then a specific metric, \emph{i.e.,} Euclidean distance, was used to perform the final classification. Recently, based on ProtoNet, an infinite mixture prototypes (IMP) network was proposed~\cite{allen2019infinite}, where each support class was represented by a set of adaptive prototypes. In addition, to avoid choosing a specific metric function, RelationNet~\cite{sung2018learning} proposed to learn a metric through a deep convolutional neural network to measure the similarity between queries and support classes.

The above methods are all based on image-level feature representations. Due to the scarcity of samples in each class in few-shot learning, the distribution of each class cannot be reliably estimated in a space of image-level features. Thus, some recent work, such as CovaMNet~\cite{li2019CovaMNet} and DN4~\cite{li2019DN4} shows that the rich local features (\emph{i.e.,} deep local descriptors) can achieve better representations than the image-level features, because the local features can be taken as a natural data augmentation operation. CovaMNet employs the second-order covariance matrix of the extracted deep local descriptors to represent each support class and designs a covariance-based metric to measure the similarities between query images and support classes. Different from CovaMNet, DN4 argues that the pooling of local features into a compact image-level representation will lose considerable discriminative information. Therefore, DN4 proposes to directly use the raw local descriptor sets to represent both query images and support classes, and then employs a cosine-based image-to-class measure to perform the relation measure.

Inspired by CovaMNet and DN4, our ADM also takes the rich deep local descriptors to represent an image. Compared with CovaMNet, the key difference is that CovaMNet only considers the distributions of the support classes but neglect the distributions of the query images, while we consider the both. Another important difference is that both CovaMNet and DN4 employ a cosine similarity function (\emph{i.e.,} an instance-level metric) to calculate a series of local relations between a query image and a certain class. In contrast, our ADM can capture the complementary global relations by using an extra distribution-level measure. In addition, we observe that the relations between query images and a certain class are actually asymmetric, \emph{i.e.,} a query image is only commensurate with an element in an image class when it is viewed as a set. Therefore, we argue that an asymmetric measure shall be considered for metric-based few-shot learning to reflect this property.


\begin{figure*}
    \centering
    \includegraphics[width=0.8\linewidth]{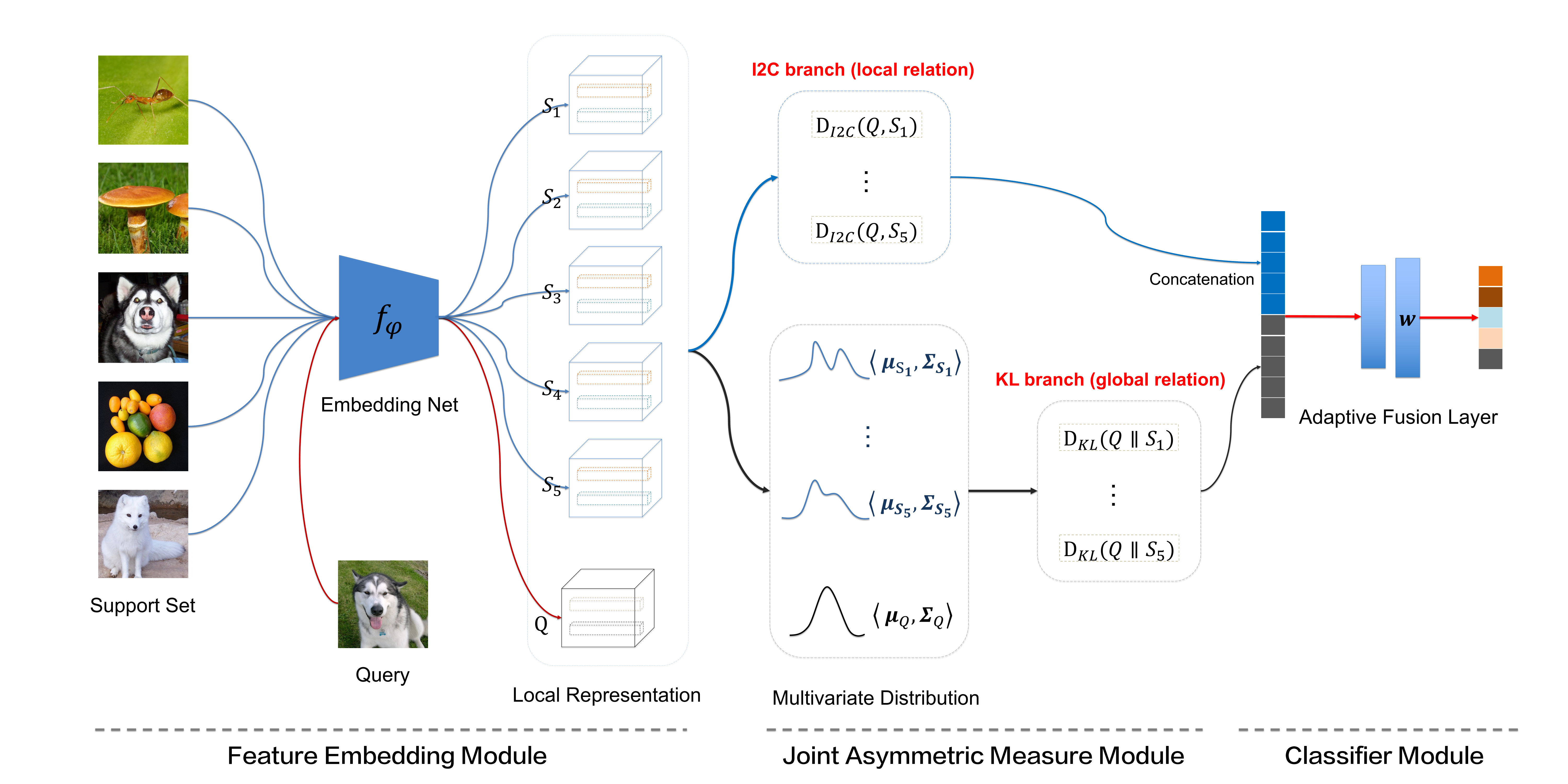}
    \caption{Architecture of the proposed Asymmetric Distribution Measure (ADM) network for a 5-way 1-shot task, which consists of three modules, \textit{i.e.,} a feature embedding module, a joint asymmetric measure module and a classifier module.}
    \label{fig:model}
\end{figure*}

\section{Preliminary}
\noindent\textbf{Problem formulation.}
Under the few-shot setting, there are usually three sets of data, \emph{i.e.,} a support set $\mathcal{S}$, a query set $\mathcal{Q}$ and an auxiliary set $\mathcal{A}$. In particular, $\mathcal{S}$ and $\mathcal{Q}$ share the same label space, which are corresponding to the training and test sets respectively in the general classification task. If $\mathcal{S}$ contains $C$ classes with $K$ (\emph{e.g.,} $1$ or $5$) samples per class, we call this classification task $C$-way $K$-shot. However, $\mathcal{S}$ only has a few samples in each class, making it almost impossible to train a deep neural network effectively. Therefore, the auxiliary set $\mathcal{A}$ is generally introduced to learn transferable knowledge to tackle this problem. Also, $\mathcal{A}$ enjoys more classes and more samples per class than $\mathcal{S}$, but has a disjoint label space from $\mathcal{S}$.

\noindent\textbf{Episodic training.}
To learn a classifier that can generalize well, an episodic training mechanism~\cite{vinyals2016matching} is normally adopted in the training stage of the metric-based few-shot learning methods. Specifically, in each episode, a new task simulating the target few-shot task is randomly constructed from $\mathcal{A}$. Each simulated task consists of two subsets, $\mathcal{A_{S}}$ and $\mathcal{A_{Q}}$, which are akin to $\mathcal{S}$ and $\mathcal{Q}$, respectively. At each iteration, one episode (task) is adopted to train the current model. Basically, tens of thousands of episodes (tasks) will be randomly sampled to train this model. Once the training process is completed, we can predict the labels of $\mathcal{Q}$ using the trained model based on $\mathcal{S}$.

\section{Methodology}

As illustrated in Figure~\ref{fig:model}, our ADM model mainly consists of three components: a feature embedding module, a joint asymmetric measure module, and a classifier module. The first module learns feature embeddings and produces rich deep local descriptors for an input image. Next, the distributions of query images and support classes can be represented at the level of deep local descriptors. The second module defines a joint asymmetric distribution measure between the query distribution and the support class distribution by considering both the asymmetric local relations and the asymmetric global relations. As for the last module, we adaptively fuse the local and global relations together by a jointly learned weight vector, and then adopt a non-parametric nearest neighbor classifier as the final classifier. These three modules are jointly trained from scratch in an end-to-end manner.

\subsection{Feature Embedding with Local Descriptors}
As have been shown by some recent work~\cite{li2019CovaMNet,li2019DN4}, local descriptor based feature representations are much richer than image-level features and can alleviate the scarcity issue of samples in few-shot learning. Following these work, we employ the rich and informative local descriptors to represent each image as well.

To this end, we design a feature embedding module $f_\varphi(\cdot)$, which can extract rich deep local descriptors for input images. Specifically, given an image $X$, $f_\varphi(X)$ will be a $c \times h \times w$ three-dimensional (3D) tensor, which can be seen as a set of $c$-dimensional local descriptors
\begin{equation}\label{function_1}
    f_\varphi(X)=[\bm{x}_1,\ldots,\bm{x}_n]\in \mathbb{R}^{c \times n}\,,
\end{equation}
where $\bm{x}_i$ is the $i$-th local descriptor and $n=h\times w$ is the total number of local descriptors for image $X$. These local descriptors can be seen as the local representations of the spatial local patches in this image. Basically, for each query image, we use the extracted $n$ local descriptors to estimate its distribution in the space of $\mathbb{R}^c$. As for each support class, all the local descriptors of all the images in this class will be used together to estimate its distribution in the space of $\mathbb{R}^c$. Since the local descriptors can capture the local subtle information, they can benefit more for the final image recognition.

\subsection{Our Asymmetric Distribution Measure (ADM)}
\label{SubSec:Model}
\paraheading{Kullback--Leibler divergence based distribution measure.}
Assuming that the distributions of local descriptors extracted from an image or a support class are multivariate Gaussian, a query image's distribution can be denoted by $Q=\mathcal{N}(\bm{\mu}_Q,\bm{\Sigma}_Q)$, and a certain support class's distribution can be expressed by $S=\mathcal{N}(\bm{\mu}_S,\bm{\Sigma}_S)$, where $\bm{\mu}\in \mathbb{R}^c$ and $\bm{\Sigma}\in \mathbb{R}^{c \times c}$ indicate the mean vector and covariance matrix of a specific distribution, respectively. Thus, Kullback-Leibler (KL) divergence~\cite{duchi2007derivations} between $Q$ and $S$ can be defined as:
\begin{equation}\label{function_2}
\begin{split}
    D_\text{KL}(Q\|S)  &= \frac{1}{2}\Big(\text{trace}(\bm{\Sigma}^{-1}_S\bm{\Sigma}_Q)+\ln\big( \frac{\det\bm{\Sigma}_S}{\det\bm{\Sigma}_Q}\big)  \\
    &+ (\bm{\mu}_S-\bm{\mu}_Q)^\top\bm{\Sigma}^{-1}_S(\bm{\mu}_S-\bm{\mu}_Q)-c \Big)\,,
\end{split}
\end{equation}
where $\text{trace}(\cdot)$ is the trace operation of matrix, $\ln(\cdot)$ denotes logarithm with the base of $e$, and $\det$ indicates the determinant of a square matrix. As seen, Eq.(\ref{function_2}) takes both the mean and covariance into account to calculate the distance between two distributions.

Typically, since the KL divergence measure is asymmetric, $D_\text{KL}(Q\|S)$ mainly matches the distribution of $Q$ to the one of $S$, which is essentially different from $D_\text{KL}(S\|Q)$. One important advantage of using Eq.(\ref{function_2}) is that it can naturally capture the asymmetric relations between query images to support classes, forcing the query images to be close to the corresponding true class when used in our network training. 


To further show the advantage of using an asymmetric measure, we purposely introduce a symmetric distribution metric function, \emph{e.g.,} $2$-Wasserstein distance~\cite{olkin1982distance}, whose formulation is defined as follows,
\begin{equation}\label{function_3}
\begin{split}
    D_\text{wass}(Q,S)^2 & =  \left\|\bm{\mu}_Q-\bm{\mu}_S\right\|_2^2+ \\
    &\text{trace}\Big(\bm{\Sigma}_Q+\bm{\Sigma}_S-2\big(\bm{\Sigma}_Q^{\frac{1}{2}}\bm{\Sigma}_S\bm{\Sigma}_Q^{\frac{1}{2}}\big)^{\frac{1}{2}}\Big)\,,
\end{split}
\end{equation}
However, due to the square root of matrices, the calculation of the above distance function is time consuming and the optimization of this function is difficult. Therefore, in the literature~\cite{berthelot2017began,he2018wasserstein}, an approximation function is normally employed
\begin{equation}\label{function_4}
    D_\text{wass}(Q,S)^2 =\left\|\bm{\mu}_Q-\bm{\mu}_S\right\|_2^2+\left\|\bm{\Sigma}_Q-\bm{\Sigma}_S\right\|_F^2\,,
\end{equation}
where the first term calculates the squared Euclidean distance between two mean vectors and the second term is a squared Frobenius norm of the difference between two covariance matrices. The comparison and analysis between $2$-Wasserstein distance and KL divergence will be detailed in Section~\ref{ablation_study}.

\paraheading{Image-to-Class based distribution measure.}
The above KL divergence measure can capture the global distribution-level relations between a query image and support classes. Nevertheless, the local relations are not taken into consideration yet. According to a deep analysis of DN4~\cite{li2019DN4}, we observe that there may be two implicit reasons of the success of DN4. One reason is that the local descriptor based measure (\emph{i.e.,} local relations) it used enjoys a stronger generalization ability than the image-level feature based measure. The other key reason is that the image-to-class measure used in DN4 is asymmetric on the whole, which aligns well with our argument of the necessity of the asymmetric measure. Therefore, such an asymmetric image-to-class measure is also introduced into our model to capture the local-level relations between queries and support classes. However, the difference in our work lies that the indispensable global relations are also complemented by an asymmetric distribution-level measure (\emph{i.e.,} KL divergence).

To be specific, given a query image $Q$ and a support class $S$, which will be represented as  $f_\varphi(Q)=[\bm{x}_1,\ldots,\bm{x}_n]\in \mathbb{R}^{c \times n}$ and $f_\varphi(S)=[f_\varphi(X_1),\ldots,f_\varphi(X_K)]\in \mathbb{R}^{c \times nK}$, respectively, where $K$ is the number of shots in $S$. Thus, the image-to-class (I2C) similarity measure can be formulated as
\begin{equation}\label{function_5}
\begin{split}
    D_\text{I2C}(Q,S) & = \sum_{i=1}^{n}\text{Topk}\Big(\frac{f_\varphi(Q)^\top\cdot f_\varphi(S)}{\|f_\varphi(Q)^\top\|_F\cdot\|f_\varphi(S)\|_F}\Big)\,,
\end{split}
\end{equation}
where $\text{Topk}(\cdot)$ means selecting the $k$ largest elements in each row of the correlation matrix between $Q$ and $S$, \emph{i.e.,} {\small$\frac{f_\varphi(Q)^\top\cdot f_\varphi(S)}{\|f_\varphi(Q)^\top\|_F\cdot\|f_\varphi(S)\|_F}$}. Typically, $k$ is set as $1$ in our work.

\paraheading{Classification with an adaptive fusion strategy.} Since two types of relations have been calculated, \textit{i.e.,} global-level relations calculated by the KL divergence measure and local-level relations produced by the I2C measure, a fusion strategy shall be designed to integrate these two parts. To tackle this issue, we adopt a learnable $2$-dimensional weight vector $\bm{w}=[w_1, w_2]$ to implement this fusion. It is worth noting that because the KL divergence indicates dissimilarity rather than similarity, we use the negative of this divergence to obtain a similarity. Specifically, the final fusion similarity between a query $Q$ and a class $S$ can be defined as follows
\begin{equation}\label{function_6}
\begin{split}
    D(Q,S) & = -w_1\cdot D_\text{KL}(Q\|S) + w_2\cdot D_\text{I2C}(Q,S)\,.
\end{split}
\end{equation}

As seen in Figure~\ref{fig:model}, for a $5$-way $1$-shot task and a specific query $Q$, the outputs of the I2C branch and KL branch are a $5$-dimensional similarity vector, respectively. Next, we concatenate these two vectors together to get a $10$-dimensional vector. And then, we apply a 1D convolution layer with the kernel size of $1\times 1$ along with a dilation value of $5$. In this way, we can obtain a weighted $5$-dimensional similarity vector by learning a $2$-dimensional weights $\bm{w}$. Additionally, a Batch Normalization layer is also added before the 1D convolution layer to balance the scale of the two parts of similarities. Finally, a non-parametric nearest neighbor classifier is performed to obtain the final classification results.

\begin{table*}[!tp]\small
\centering
\tabcolsep=6pt
\caption{Ablation study on both \textit{mini}ImageNet and \textit{tiered}ImageNet. The second column refers to whether the measure function adopted is symmetric or not. The third column indicates which kind of measure function is employed, \emph{i.e.,} instance-level or distribution-level. For each setting, the best and the second best methods are highlighted.}
\label{Ablation_Study}
\begin{tabular}{l c  c c c c c}
\toprule[1pt]
\multirow{2}{*}{\textbf{Method}} & \multirow{2}{*}{\textbf{Type}} & \multirow{2}{*}{\textbf{Measure}}   & \multicolumn{2}{c}{\textbf{\textit{mini}ImageNet 5-way Acc (\%)}}  & \multicolumn{2}{c}{\textbf{\textit{tiered}ImageNet 5-way Acc (\%)}}\\ 
\cmidrule{4-7} 
&   & &1-shot &5-shot  &1-shot &5-shot\\ 
\midrule
\textbf{ProtoNet}$^\ddag$ [NeurIPS 2017]  & Symmetric & Instance-level  &\myacc{48.45}{0.96}   &\myacc{66.53}{0.51}  &\myacc{48.58}{0.87}  &\myacc{69.57}{0.75}\\
\textbf{RelationNet} [CVPR 2018]          & Symmetric & Instance-level  &\myacc{50.44}{0.82}   &\myacc{65.32}{0.70}  &\myacc{54.48}{0.93}  &\myacc{71.31}{0.78}\\
\midrule
\textbf{Wasserstein} (Ours)                  & Symmetric & Distribution-level  &\myacc{50.27}{0.62}   &\myacc{67.50}{0.52}  &\myacc{52.76}{0.71}  &\myacc{73.58}{0.57}\\
\textbf{Wass-CMS} (Ours)                     & Symmetric  & Distribution-level &\myacc{50.80}{0.64}   &\myacc{68.36}{0.50}  &\myacc{53.48}{0.68}  &\myacc{73.95}{0.56}\\
\textbf{KL} (Ours)                           & Asymmetric & Distribution-level &\maxacc{52.94}{0.63}  &\maxacc{69.38}{0.51} &\maxacc{55.59}{0.70} &\maxacc{74.21}{0.56}\\
\textbf{KL-CMS} (Ours)                       & Asymmetric & Distribution-level &\maxacc{53.10}{0.62}  &\maxacc{69.73}{0.50} &\maxacc{56.54}{0.70} &\maxacc{74.83}{0.56}\\
\bottomrule
\end{tabular}   
\end{table*}

\subsection{Our Contrastive Measure Strategy (CMS)}
To make the distribution measure more discriminative, we further propose an alternative task-aware \textit{Contrastive Measure Strategy (CMS)} by introducing additional contrastive information. Specifically, for a specific support set $\mathcal{S}=\{S_1,\cdots,S_C\}$, where $C$ is the number of classes in $\mathcal{S}$, we construct a \textit{distribution-level triplet} $\langle Q,S_i,S_i'\rangle$. In this triplet, $Q$ denotes a query distribution, $S_i$ is one class distribution we want to match $Q$ with, and $S_i'$ indicates the entire distribution of the remaining classes $S_j|_{j=1}^C(j\neq i)$. In this way, we can define the contrastive KL divergence measure as follows
\begin{equation}
        D_\text{KL}^\text{con}(Q\| S_i) = D_\text{KL}(Q\| S_i) - D_\text{KL}(Q\|S_i')\,.
\end{equation}

The advantage of using the above contrastive measure function over merely using $D_\text{KL}(Q\|S_i)$ in Eq.(\ref{function_2}) is that the context of the entire support classes is taken into consideration. In this way, we can take a whole view of the entire task when measuring the relation between $Q$ and each individual class $S_i$, making the measure function more discriminative. This will be experimentally demonstrated shortly.

\section{Experiments}
In this section, extensive experiments on two benchmark datasets are conducted, including an ablation study.

\subsection{Datasets}
All experiments are conducted on two popular few-shot learning benchmarks, \emph{i.e.,} \textit{mini}ImageNet \cite{vinyals2016matching} and \textit{tiered}ImageNet \cite{ren2018meta}.

\textbf{\textit{mini}ImageNet.} This dataset is widely used in few-shot learning, which is a small subset of ImageNet~\cite{deng2009imagenet}. It contains $100$ classes with $600$ images in each class. We use the same splits as in~\cite{ravi2016optimization}, which takes $64$, $16$ and $20$ classes for training, validation and test, respectively.

\textbf{\textit{tiered}ImageNet.} Similar to \textit{mini}ImageNet, it is also a mini-version of ImageNet. However, they are different in two aspects. The first is that \textit{tiered}ImageNet has a larger number of classes ($608$ classes) and more images for each class ($1281$ images per class). The other difference is that \textit{tiered}ImageNet has a hierarchical structure of categories. Specifically, there are $34$ categories at the top hierarchy and they are split into $20$ training categories ($351$ classes), $6$ validation categories ($97$ classes) and $8$ test categories ($160$ classes). On this dataset, we strictly follow the splits used in~\cite{ren2018meta}.

For both \textit{mini}ImageNet and \textit{tiered}ImageNet, the resolution of all the images is resized to $84\times 84$.

\begin{table*}[!tp]\small
\tabcolsep=5.5pt
\centering
\caption{The mean accuracies of the 5-way 1-shot and 5-shot tasks on both \textit{mini}ImageNet and \textit{tiered}ImageNet, with 95\% confidence intervals. The third column refers to which kind of embedding network is employed. The fifth column shows the total parameters used by each method. $^\ddag$ Results are obtained by the re-implemented version in the same setting. For each setting, the best and the second best methods are highlighted.}
\label{Table2_SOTA}
\begin{tabular}{l c c c c c c c c}
\toprule[1pt]
\multirow{2}{*}{\textbf{Method}} & \multirow{2}{*}{\textbf{Venue}} & \multirow{2}{*}{\textbf{Embed.}} & \multirow{2}{*}{\textbf{Type}} & \multirow{2}{*}{\textbf{Para.}} & \multicolumn{2}{c}{\textbf{\textit{mini}ImageNet 5-way Acc (\%)}}  & \multicolumn{2}{c}{\textbf{\textit{tiered}ImageNet 5-way Acc (\%)}} \\ \cmidrule{6-9}
& & & & & 1-shot & 5-shot & 1-shot & 5-shot \\ 
\midrule
Meta LSTM          & ICLR'17                 & \textit{Conv-32F}     & Meta   & -      & \myacc{43.44}{0.77}   & \myacc{60.60}{0.71}   & -   & -\\
MAML               & ICML'17                 & \textit{Conv-32F}     & Meta   & -      & \myacc{48.70}{1.84}   & \myacc{63.11}{0.92}   & \myacc{51.67}{1.81}  & \myacc{70.30}{1.75}\\
SNAIL              & ICLR'18                 & \textit{Conv-32F}     & Meta   & -      & 45.10                 & 55.20                 & -   & - \\
MTL                & CVPR'19                 & \textit{Conv-32F}     & Meta   & -      & \myacc{45.60}{1.80}   & \myacc{61.20}{0.90}   & -   & -\\
TAML-Entropy       & CVPR'19                 & \textit{Conv-32F}     & Meta   & -      & \myacc{49.33}{1.80}   & \myacc{66.05}{0.85}   & -   & -\\
MetaOptNet-RR      & CVPR'19                 & \textit{Conv-64F}     & Meta   & -      & \myacc{52.87}{0.57}   & \myacc{69.51}{0.48}   & \myacc{54.63}{0.67}   & \myacc{72.11}{0.59}\\
\cmidrule{1-9}
\textbf{Matching Nets}      & NeurIPS'16     & \textit{Conv-64F}     & Metric & 113 kB   & \myacc{43.56}{0.84}   & \myacc{55.31}{0.73}   & -   & - \\
\textbf{ProtoNet}$^\ddag$   & NeurIPS'17     & \textit{Conv-64F}     & Metric & 113 kB   & \myacc{48.45}{0.96}   & \myacc{66.53}{0.51}   & \myacc{48.58}{0.87}   & \myacc{69.57}{0.75} \\
\textbf{RelationNet}        & CVPR'18        & \textit{Conv-64F}     & Metric & 228 kB   & \myacc{50.44}{0.82}   & \myacc{65.32}{0.70}   & \myacc{54.48}{0.93}   & \myacc{71.31}{0.78}\\
\textbf{IMP}                & ICML'19        & \textit{Conv-64F}     & Metric & 113 kB   & \myacc{49.6}{0.8}     & \myacc{68.1}{0.8}     & -   & -\\
\textbf{CovaMNet}           & AAAI'19        & \textit{Conv-64F}     & Metric & 113 kB   & \myacc{51.19}{0.76}   & \myacc{67.65}{0.63}   & \myacc{54.98}{0.90}   & \myacc{71.51}{0.75} \\
\textbf{DN4}                & CVPR'19        & \textit{Conv-64F}     & Metric & 113 kB   & \myacc{51.24}{0.74}   & \maxacc{71.02}{0.64}  & \myacc{53.37}{0.86}   & \myacc{74.45}{0.70} \\
\cmidrule{1-9}
\textbf{KL}                 & Ours           & \textit{Conv-64F}     & Metric & 113 kB   &\myacc{52.94}{0.63}    &\myacc{69.38}{0.51}    &\myacc{55.59}{0.70}     &\myacc{74.21}{0.56}\\
\textbf{KL-CMS}             & Ours           & \textit{Conv-64F}     & Metric & 113 kB   & \maxacc{53.10}{0.62}  & \myacc{69.73}{0.50}   & \maxacc{56.54}{0.70}   & \maxacc{74.83}{0.56}\\
\textbf{ADM}                & Ours           & \textit{Conv-64F}     & Metric & 113 kB   & \maxacc{54.26}{0.63}  & \maxacc{72.54}{0.50}  & \maxacc{56.01}{0.69}   & \maxacc{75.18}{0.56}\\
\bottomrule
\end{tabular}   
\end{table*}

\subsection{Network Architecture}
It can be easily verified that adopting a deeper network for embedding or using pre-trained weights will provide higher accuracy. Following the previous works~\cite{snell2017prototypical,sung2018learning,li2019CovaMNet,li2019DN4}, we adopt the same embedding network with four convolutional blocks, \textit{i.e.,} \textit{Conv-64F}, to make a fair comparison with other methods. Specifically, the first two blocks each contains a convolutional layer (with $64$ filters of size $3\times 3$), a batch-normalization layer, a Leaky ReLU layer and a max pooling layer. The last two blocks adopt the same architecture but without pooling layers. The reason for only using two pooling layers is that we need richer local descriptors to represent the distributions of both queries and classes. For example, in a $5$-way $1$-shot setting, when the size of the input image is $84 \times 84$, we can only obtain $25$ local descriptors for each image (class) by adopting four pooling layers. It is clearly insufficient to represent a distribution with a feature dimensionality of $64$. In contrast, using the adopted network architecture with two pooling layers, we obtain $441$ local descriptors for each image (class).

\subsection{Implementation Details}
Both $5$-way $1$-shot and $5$-way $5$-shot classification tasks are conducted to evaluate our methods. We use $15$ query images per class in each single task ($75$ query images in total) in both training and test stages. In particular, we employ the episodic training mechanism~\cite{vinyals2016matching} to train our models from scratch without pre-training. In the training stage, we use the Adam algorithm~\cite{kingma2014adam} to train all the models for $40$ epoches. In each epoch, we randomly construct $10000$ episodes (tasks). Also, the initial learning rate is set as $1 \times 10^{-3}$ and multiplied by $0.5$ every $10$ epoches. During test, $1000$ tasks are randomly constructed to calculate the final results, and this process is repeated five times. The top-1 mean accuracy is taken as the evaluation criterion. At the same time, the $95\%$ confidence intervals are also reported.

\subsection{Comparison Methods}
Since our methods belong to the metric-based few-shot learning methods, we will mainly compare our methods with metric-based methods, such as Matching Net~\cite{vinyals2016matching}, ProtoNet~\cite{snell2017prototypical}, RelationNet~\cite{sung2018learning}, IMP~\cite{allen2019infinite}, CovaMNet~\cite{li2019CovaMNet} and DN4~\cite{li2019DN4}. Moreover, representative meta-learning based few-shot learning methods are also listed for reference, including Meta LSTM~\cite{ravi2016optimization}, MAML~\cite{finn2017model}, SNAIL~\cite{mishra2017simple}, MTL~\cite{sun2019meta}, TAML-Entropy~\cite{jamal2019task}, and MetaOptNet-RR~\cite{lee2019meta}. Note that meta-learning based methods are essentially different from metric-based methods at two aspects. The first aspect is that an additional parameterized meta-learner is usually learned in meta-learning based methods while the metric-based methods do not have. The second aspect is that during test, meta-learning based methods will fine-tune the model (or classifier) to obtain the final classification results while metric-based methods do not need fine-tuning.

Most results of these compared methods are quoted from their original work or the relevant reference. Some methods are not in the same setting with our method, such as ProtoNet, so we use the results of their modified versions to ensure fair comparison. For some recent meta-based methods, such as SNAIL, MTL and TAML-Entropy, we only report their results with a similar embedding network, \emph{e.g.,} \textit{Conv-32F}, which has the same architecture with \textit{Conv-64F} but has $32$ filters in each convolutional block.

\subsection{Ablation Study}
\label{ablation_study}
In this section, we first verify the validity of our argument on asymmetric measure for metric-based few-shot learning. Next, based on two distribution-level measure functions, we evaluate the effectiveness of the proposed CMS strategy. Specifically, both the $2$-Wasserstein distance (Wasserstein for short) and KL divergence (KL for short) are performed on the \textit{mini}ImageNet and \textit{tiered}ImageNet datasets. Also, the contrastive versions using our proposed CMS are named as Wass-CMS and KL-CMS, respectively. Moreover, two instance-level symmetric metric based methods, \textit{i.e.,} ProtoNet and RelationNet, are picked as baselines.

As seen in Table~\ref{Ablation_Study}, compared to symmetric metric based methods, such as ProtoNet, RelationNet and Wasserstein, the proposed asymmetric measure can obtain superior results. For example, on the \textit{mini}ImageNet, KL gains $4.49\%$, $2.50\%$ and $2.67\%$ over these methods on the $1$-shot task, respectively. This verifies that an asymmetric measure is more suitable for metric-based few-shot learning.

We can also see that the proposed CMS strategy can indeed improve the performance of distribution-based measure functions, especially on the $1$-shot setting. For instance, on the \textit{tiered}ImageNet, Wass-CMS achieves $0.72\%$ improvement over Wass, and KL-CMS obtains $0.95\%$ improvement over KL on the $1$-shot task. This shows that the task-aware CMS strategy does enhance the distribution-based measure functions, thanks to taking a whole view of the entire task.

\subsection{Comparison with the State of the Art}
Experimental results on the comparison with the state-of-the-art methods are reported in Table~\ref{Table2_SOTA}, where two types of few-shot learning methods (\emph{i.e.,} both meta-learning based and metric-based) are compared. Since our methods are metric-based methods, we will mainly compare our methods with other metric-based ones. Moreover, the total number of parameters of each method is also shown in the fifth column.

From Table~\ref{Table2_SOTA}, it can be seen that the proposed ADM (without CMS) outperforms all the other metric-based and meta-learning based methods on both $1$-shot and $5$-shot settings. For example, on the \textit{mini}ImageNet, our ADM obtains $10.7\%$, $5.81\%$, $3.82\%$, $4.66\%$, $3.07\%$ and $3.02\%$ improvements over Matching Nets, ProtoNet, RelationNet, IMP, CovaMNet and DN4 on the $1$-shot task, respectively. Moreover, on the \textit{tiered}ImageNet, our ADM achieves $5.61\%$, $3.87\%$, $3.67\%$, $0.73\%$ improvements over ProtoNet, RelationNet, CovaMNet and DN4 on the $5$-shot task, respectively. This verifies the effectiveness and superiority of our proposed ADM, owing to the integration of both local and global asymmetric relations.

The proposed KL and KL-CMS are also very competitive with the state-of-the-art methods. Specifically, on the $1$-shot setting, KL and KL-CMS can obtain significantly improvements over the existing metric-based methods. For instance, on the \textit{mini}ImageNet, KL/KL-CMS gains $9.38\%/9.54\%$, $4.49\%/4.65\%$, $2.5\%/2.66\%$, $3.34\%/3.5\%$, $1.75\%/1.91\%$ and $1.7\%/1.86\%$ improvements over Matching Nets, ProtoNet, RelationNet, IMP, CovaMNet and DN4, respectively. It verifies that such kind of distribution-based asymmetric measure is more suitable for metric-based few-shot learning.

\section{Conclusion}
In this study, we provide a new perspective for metric-based few-shot learning by considering the asymmetric nature of the similarity measure and design a novel \textit{Asymmetric Distribution Measure (ADM)} network to address this task. Furthermore, to make full use of the context of the entire task, we propose a \textit{Contrastive Measure Strategy (CMS)} to learn a more discriminative distribution metric space. Extensive experiments on two benchmark datasets verify the effectiveness and advantages of both local asymmetric relations and global asymmetric relations in metric-based few-shot learning.

\bibliographystyle{named}
\bibliography{ijcai20}

\end{document}